\documentclass[10pt,twocolumn,letterpaper]{article}

 \usepackage[pagenumbers]{cvpr} 

\usepackage{booktabs}
\usepackage{multirow}
\usepackage{makecell}
\usepackage{amssymb}
\usepackage{amsmath}
\usepackage{algorithm2e}
\RestyleAlgo{ruled}
\usepackage{xcolor}
\usepackage{pifont}
\definecolor{cvprblue}{rgb}{0.21,0.49,0.74}
\usepackage[pagebackref,breaklinks,colorlinks,allcolors=cvprblue]{hyperref}

\def\paperID{*****} 
\def\confName{CVPR}
\def\confYear{2026}

\title{SyncEdit: Rethinking Lip Synchronization as Editing with Audio-Driven Diffusion Models}

\author{Lixiang Lin\\
HiThink Research\\
{\tt\small linlixiang@myhexin.com}
\and
Siyuan Jin\\
USTC\\
\and
Jinshan Zhang\\
ZheJiang University
}

\begin{document}
\maketitle
\begin{abstract}
Lip synchronization refers to the task of modifying facial lip movements such that they are temporally aligned with a given audio signal. It is a fundamental problem in audio-visual synthesis for generating realistic talking-head videos. Recent progress in diffusion-based generative models has led to remarkable advances in lip synchronization. Nevertheless, existing methods typically achieve audio-visual alignment by fine-tuning pre-trained diffusion models on large scale datasets, resulting in substantial computational costs and significant data requirements. Inspired by FlowEdit, we reformulate lip synchronization as a video editing problem. Building upon a pre-trained audio-driven diffusion model, our approach achieves lip synchronization in a training-free manner, without additional fine-tuning or paired data. In this paper, we present SyncEdit, a training-free framework designed for lip synchronization. We reformulate the editing paradigm by substituting the edit sequence in FlowEdit with the target sequence, yielding an unbiased estimation of the desired output. Moreover, we propose annealed noise alignment, which progressively alignes the sampled Gaussian noise with diffusion-model-estimated noise during iterative editing, producing a smooth and stable editing trajectory. Extensive experimental results validate the effectiveness and robustness of the proposed framework. Code is available at \href[]{https://github.com/l1346792580123/SyncEdit}{here}.
\end{abstract}    
\section{Introduction}
\label{sec:intro}

Lip synchronization~\cite{PrajwalMNJ20wav2lip,Zhong2023iplap,li2024latentsync}, which aims to align mouth movements with corresponding speech audio, is a fundamental yet critical task in computer vision. By enforcing fine-grained temporal and semantic consistency between facial dynamics and acoustic signals, lip synchronization plays an essential role in applications such as film dubbing and digital human generation. As a result, it has become a foundational component in modern video generation systems, underpinning the realism and perceptual quality of AI-generated visual content.

Recent advances in diffusion-based generative models~\cite{HoJA20ddpm,LipmanCBNL23flowmatching,RombachBLEO22sd,PodellELBDMPR24sdxl} have significantly accelerated the development of lip synchronization techniques. Numerous lip synchronization approaches~\cite{Mukhopadhyay2024diff2lip, Zhang2024musetalk, peng2025omnisync} formulate lip synchronization as a conditional generation problem and achieve impressive performance by fine-tuning text-to-image or text-to-video diffusion models. However, their success typically relies on large amounts of paired audio-visual training data and substantial computational resources for model adaptation. These limitations motivate the exploration of training-free alternatives that can exploit the powerful generative capabilities of pretrained diffusion models while avoiding costly fine-tuning procedures.

\begin{figure}[!t]
	\centering
	\includegraphics[width=\linewidth]{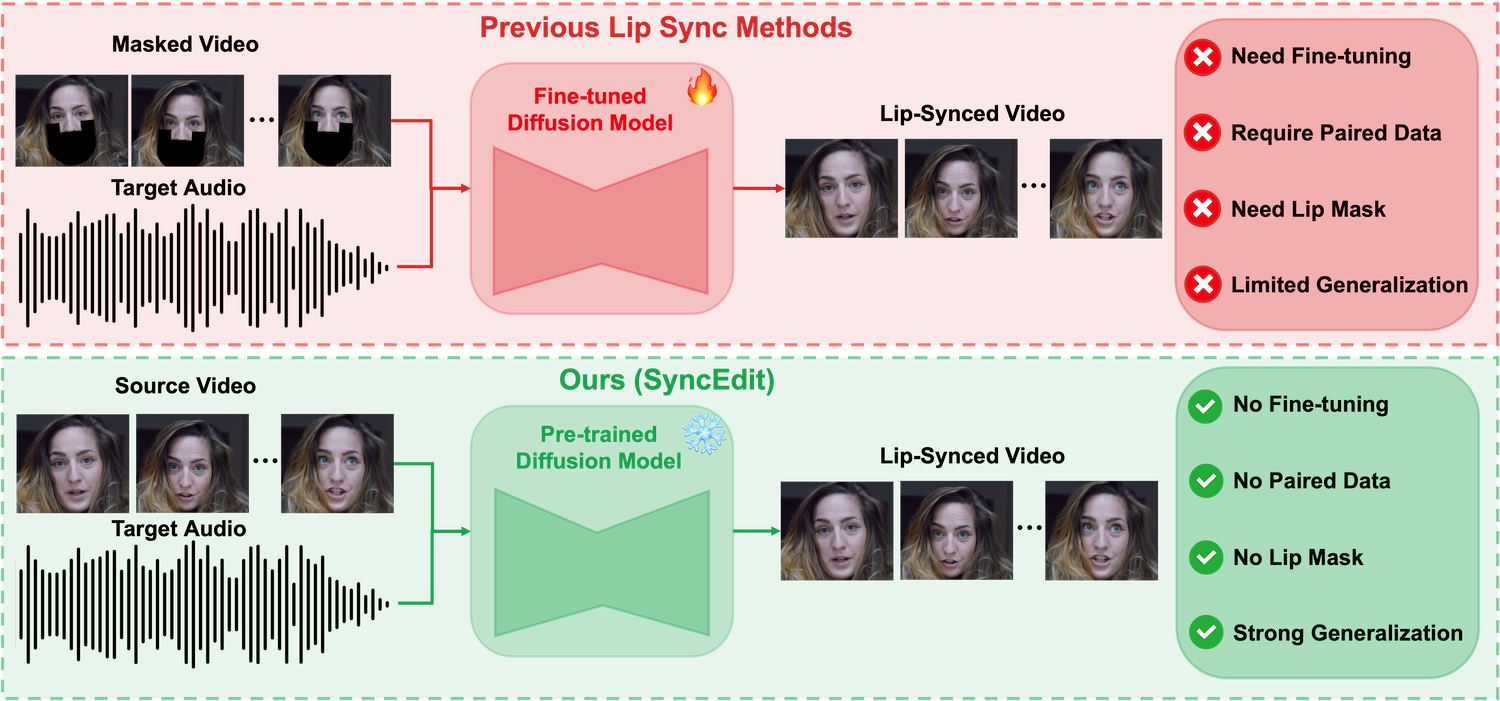}
	\caption{\textbf{Comparison of existing lip synchronization methods and our approach}. Previous methods generally rely on paired audio–video data, diffusion model fine-tuning, and lip masking, whereas our method achieves lip synchronization using only a pretrained audio-driven diffusion model. }
	\label{fig:cmp}
\end{figure}

Moreover, due to the scarcity of paired lip-synchronization training data, many existing methods rely on explicit mouth-region masking or employ carefully designed training strategies to enforce audio-visual alignment. Such designs not only increase data  annotation requirements but also introduce strong assumptions about facial geometry. Consequently, their performance often degrades in challenging scenarios, such as facial occlusions, large head poses, and profile views.

Motivated by the aforementioned limitations, we revisit lip synchronization from a different perspective by formulating it as a video editing problem. We propose SyncEdit, a training-free framework that directly leverages the rich audio-visual priors of pre-trained audio-driven diffusion models. As illustrated in Fig.~\ref{fig:cmp}, unlike existing diffusion-based approaches that require task-specific fine-tuning and paired lip-sync data, SyncEdit performs lip synchronization by directly editing the input video according to the target speech, without any additional training. By treating lip synchronization as an editing process, our method achieves accurate temporal alignment between lip movements and speech while preserving the speaker's identity and non-lip facial attributes. SyncEdit substantially reduces both data and computational requirements, while naturally supporting flexible audio-driven visual editing across a wide range of scenarios.

We first replace the iterative refinement of the edit sequence in FlowEdit~\cite{kulikov2025flowedit} with iterations over the target sequence, obtaining an unbiased estimation of the desired output. This reformulation not only reduces the inherent bias introduced by sequential editing but also enables a more direct alignment with the target distribution.

Second, we replace purely stochastic Gaussian sampling with an annealed noise alignment strategy, where Gaussian noise is progressively blended with noise estimated by the pre-trained diffusion model. As the iterative process proceeds, the contribution of Gaussian noise gradually diminishes, resulting in a smoother optimization trajectory that improves both the stability and quality of the generated results

Our contributions can be summarized as follows:

\begin{itemize}
	\item We present \textbf{SyncEdit}, the first training-free framework for lip synchronization, eliminating the need for task-specific fine-tuning, and paving the way for plug-and-play multimodal content creation.
	\item We establish a target iterative sequence to enable an unbiased estimation of the desired output, reducing bias from sequential editing and facilitating more direct alignment with the target distribution.
	\item We construct a smoother and more stable generation trajectory through annealed noise alignment, where stochastic Gaussian noise is progressively aligned with noise estimated by a pre-trained diffusion model, leading to improved generation stability and output quality.
	\item SyncEdit achieves state-of-the-art performance without additional training, rivaling or surpassing supervised methods in lip synchronization while supporting flexible cross-modal audio–visual editing.
\end{itemize}

\begin{figure}[!t]
	\centering
	\includegraphics[width=\linewidth]{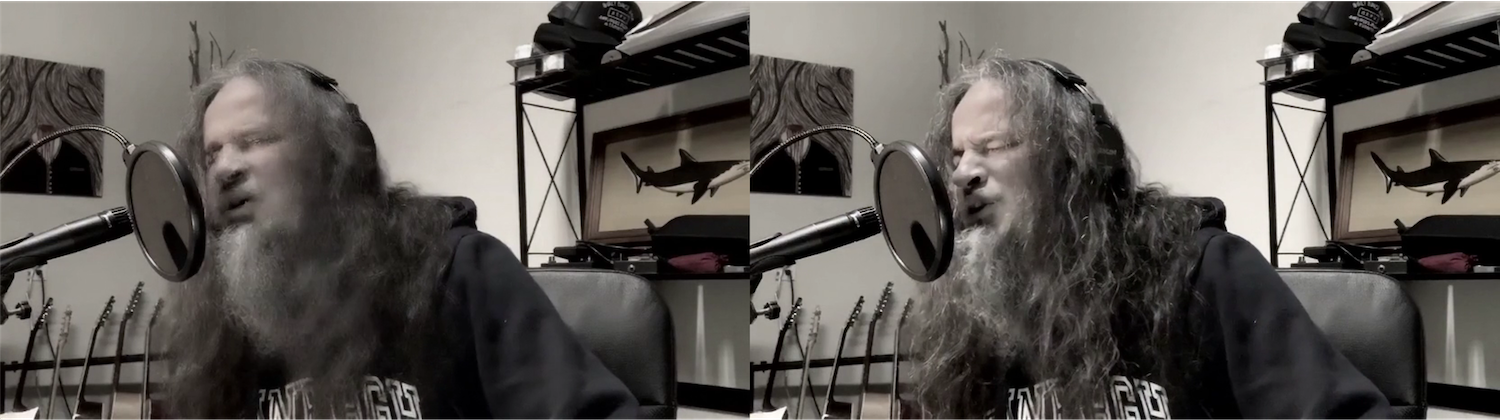}
	\caption{\textbf{Comparison of editing results under different $t_{max}$ values}. The left image presents the result obtained with $t_{max}=1$, whereas the right image corresponds to $t_{max}=0.95$. Starting the editing process from pure noise ($t_{max} = 1$) degrades the fidelity of the edited image}
	\label{fig:int1}
\end{figure}
\section{Related Works}
\label{sec:relat}

\subsection{Audio-driven Video Generation}


Audio-driven video generation is commonly categorized into lip synchronization and portrait animation, distinguished primarily by their input settings. Lip synchronization operates on a given video to match target audio, while portrait animation generates video content using only audio and a reference image. The emergence of diffusion models has markedly advanced the state of the art in audio-driven lip synchronization~\cite{li2024latentsync, Zhang2024musetalk,ma25sayanything,peng2025omnisync}. Omnisync~\cite{peng2025omnisync} adopts a mask-free training strategy with diffusion transformer models, allowing direct frame-level editing in the absence of explicit masking. Despite achieving promising results, these approaches still requires paired audio-visual data and additional model fine-tuning.

Portrait Animation~\cite{TianWZB24emo, meng2025echomimicv3, LYZSCZZ025hallo2, JiHXZLHZ0CLLW25sonic, chen2025humo,gan2025omniavatar,gao2025wans2v} differs from lip synchronization in that it directly generates videos by driving a reference image with an audio signal. While this approach requires fewer conditional inputs, it typically offers limited control over background, head pose, and detailed facial expressions. We propose a training-free method that leverages a portrait animation model to achieve lip synchronization, hence unifying the two approaches.

\begin{figure*}[!t]
	\centering
	\includegraphics[width=\linewidth]{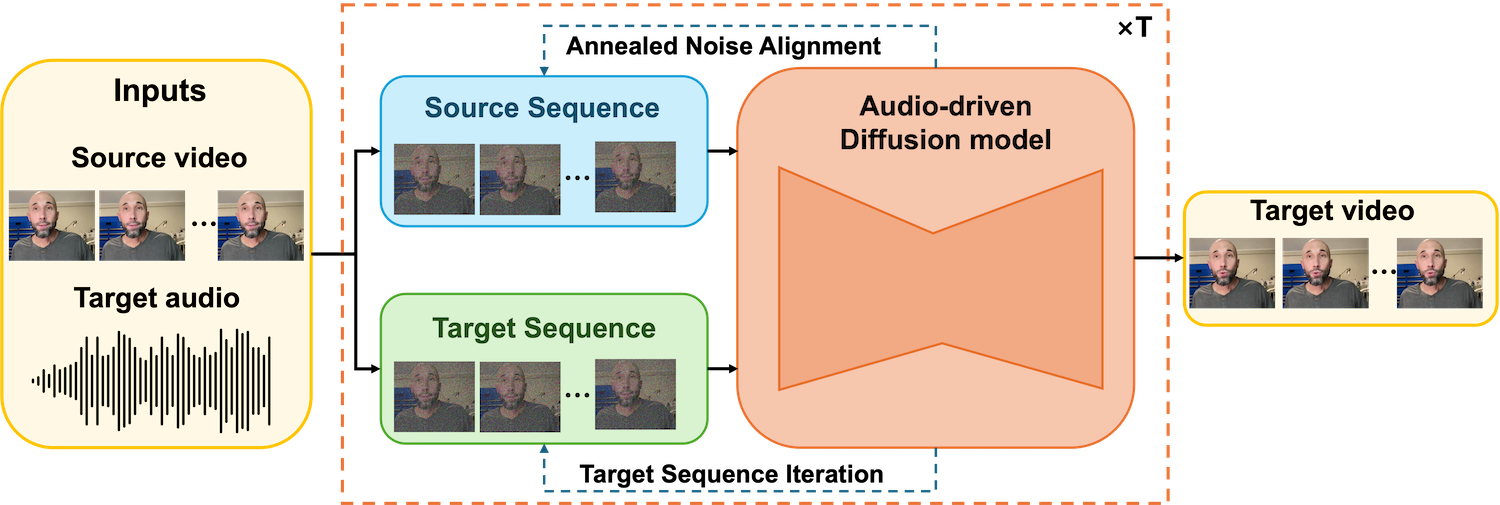}
	\caption{\textbf{Overview of SyncEdit}. Given a source video and a target audio, SyncEdit performs lip synchronization through a pre-trained audio-driven diffusion model. SyncEdit directly iterates over the target sequence, yielding an unbiased estimate of the desired output. Furthermore, SyncEdit employs an annealed noise alignment strategy that progressively replaces stochastic Gaussian noise with model-estimated noise throughout the denoising process.}
	\label{fig:overview}
\end{figure*}

\subsection{Training Free Editing}

Training-free editing has emerged as an important direction in diffusion-based generative modeling, as it enables image manipulation without requiring additional model retraining or task-specific fine-tuning. Optimization-based editing~\cite{KimPY24dreamsampler, KooPS24posteriorsampling, NamKPY24constrastive, HertzAC23deltadenoise} directly modifies the input image by iteratively optimizing it toward consistency with a given textual prompt. Inversion-based editing~\cite{HertzMTAPC23p2p, SamuelMMTDACB25lightning,wallace2022edict,Tumanyan_2023_plug} involves recovering a noise map via image-to-noise inversion and subsequently applying the conventional diffusion process guided by the target prompts to achieve the intended modifications.

FlowEdit~\cite{kulikov2025flowedit} introduces an inversion-free, optimization-free and model agnostic method for text-based image editing. FlowEdit constructs a direct ODE between the source and target distributions, without passing through the standard Gaussian distribution. FlowAlign~\cite{Kim2025flowalign} extends FlowEdit by promoting a smoother transition from source to target. FlowDirector~\cite{Li2025flowdirector} further introduces flow correction and applies the framework to text-to-video editing.
\section{Preliminaries}
\label{sec:preli}

\subsection{Flow Matching}

Flow Matching~\cite{LipmanCBNL23flowmatching} is a framework for learning continuous-time generative models by directly regressing the vector field of a probability flow. The transportation between the data distribution $X_0$ and the Gaussian distribution $X_1 \sim \mathcal{N}(0,I)$, is defined  by an ordinary differential equation (ODE)
\begin{equation}
	dX_t = V(X_t, t) dt,
\end{equation}
$t \in [0,1]$, $V$ is the velocity field. Flow matching trains a neural network $v_{\theta}$ to model the velocity field by minimizing
\begin{equation}
	\mathcal{L}_{FM} = \mathbb{E} || v_{\theta}(x_t,t) - V(x_t,t| x_1) ||^2,
\end{equation}
where $x_t \sim X_t, x_1 \sim X_1$, $V(x_t,t|x_1)$ is the conditional velocity field. Once $v_{\theta}$ is trained, the data sample can be generated by sampling a random Gaussian noise and solving the ODE iteratively.

Rectified Flow~\cite{LiuG023rectifiedflow} is a particular instantiation of flow matching that constructs a linear interpolation path between samples from the data distribution and the Gaussian distribution $x_t = (1-t)x_0 + tx_1$, resulting in a constant target velocity $V(x_t,t|x_1) = x_1 - x_0$ that transports noise to data along straight-line trajectories.

For conditional generation, the velocity field is parameterized to depend on auxiliary conditioning information, such as text or audio. Conditional generative flow models are trained on paired observations of data and conditions, enabling sampling from the corresponding conditional data distribution.

\subsection{FlowEdit}

FlowEdit~\cite{kulikov2025flowedit} presents an inversion-free and model agnostic method for text-based image editing using pretrained flow models. FlowEdit reinterprets inversion as a direct path between the source and target distribution, and defines an edit sequence

\begin{align}
	X_t^{edit} &= X^{src} + X_t^{tar} - X_t^{src}, \label{eq:flowedit} \\
	X_t^{src} &= (1-t)X^{src} + t \epsilon.
\end{align}
$X^{src}$ is the source image, $\epsilon \sim \mathcal{N}(0,I)$ is Gaussian noise, $X_t^{src}$ and $X_t^{tar}$ represent the sequence of source image and target image respectively. $X_t^{src}$ is constructed by adding Gaussian noise to the source image, and $X_t^{tar}$ is computed according to Eq.~\ref{eq:flowedit} $X_t^{tar} = X_t^{edit} + X_t^{src} - X^{src}$. The transportation ODE of $X_t^{edit}$ is defined by
\begin{equation}
	dX_t^{edit} = (V(X_t^{tar},t) - V(X_t^{src}, t)) dt. \label{eq:editode}
\end{equation}
For more details, we suggest readers to refer to the FlowEdit~\cite{kulikov2025flowedit} paper.

While FlowEdit defines the direct transportation from source image to target image, making the iteration results noise-free, some problems still exist. First, the editing process starts from $t_{max}$ instead of $1$, and $X_{t_{max}}^{edit}$ is initialized to $X^{src}$. Consequently, the resulting edited output is not strictly equivalent to the desired target sample $X^{tar}$ Instead, it is given by
\begin{equation}
	X^{edit} = X^{src} + \int_{t_{max}}^{0}(V(X_t^{tar},t) - V(X_t^{src}, t)) dt,
\end{equation}
which reveals that the final estimate depends on the discrepancy between the target and source velocity fields accumulated along the truncated trajectory. This mismatch between the initialization scheme and the ideal diffusion boundary condition introduces an inherent bias in the editing process, preventing exact recovery of the target distribution. 

\begin{figure}[!t]
	\centering
	\includegraphics[width=\linewidth]{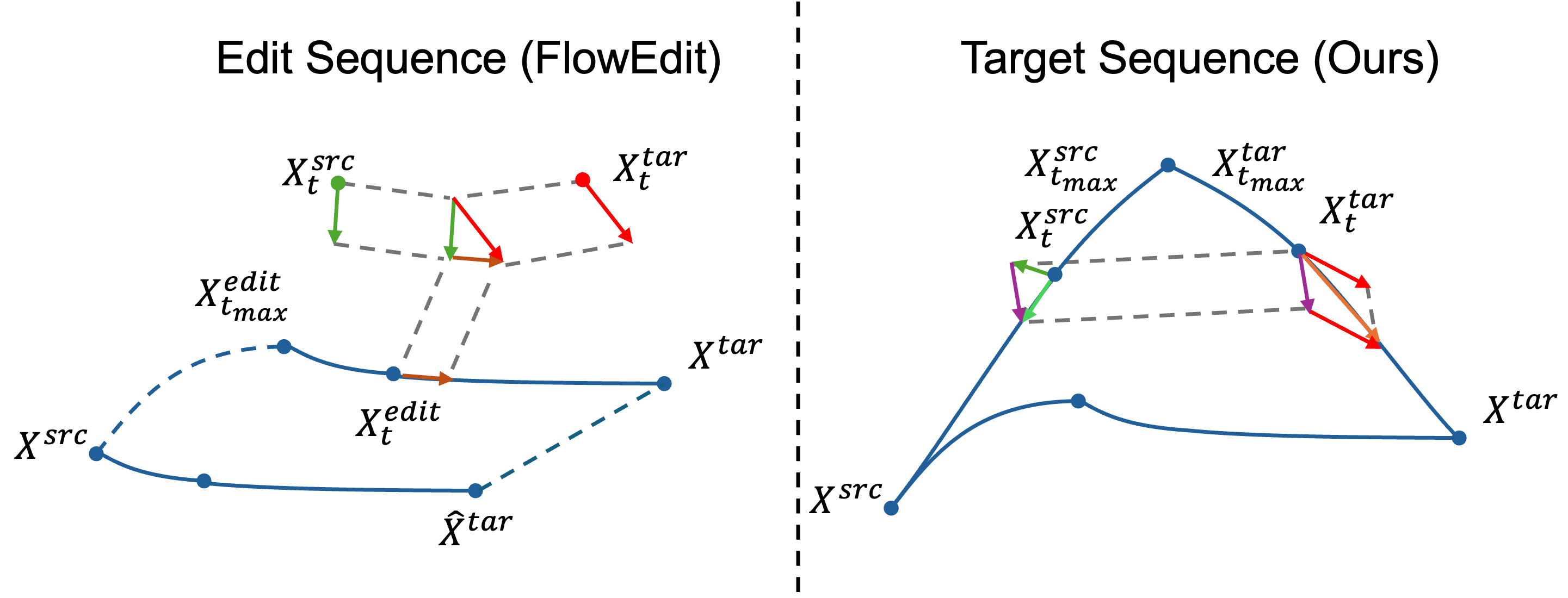}
	\caption{\textbf{Illustration of the difference between edit sequence and the proposed target-sequence iteration}. FlowEdit estimates the edited sample by integrating an edit sequence initialized at $t_{max}$, which introduces estimation bias with respect to the desired target. In contrast, the proposed target-sequence formulation directly iterates over the target trajectory, eliminating the bias and producing an unbiased estimate of the final output.}
	\label{fig:traj}
\end{figure}

Although initializing the trajectory from $t=1$ is more faithful to the original diffusion process, it incurs higher computational overhead by integrating over the entire time horizon. In addition, initiating the editing process from pure Gaussian noise amplifies numerical error accumulation and trajectory instability, often leading to degraded editing fidelity. Fig~\ref{fig:int1} demonstrate the impact of different choices of $t_{max}$.

Second, as noted in FlowAlign~\cite{Kim2025flowalign} and DynaEdit~\cite{kulikov2026dynaedit}, sampling a random Gaussian noise $\epsilon_t$ at each iteration to generate $x_t^{src}$ produces a non-smooth trajectory. This temporal inconsistency propagates to $x_t^{edit}$, reducing optimization stability,  and hindering the attainment of optimal performance.

\begin{algorithm}[!t]
	\SetKwInput{KwInit}{Init}
	\SetKwInput{KwReturn}{Return}
	\newcommand{\mycommentfont}[1]{\textcolor{blue}{#1}}
	\SetCommentSty{mycommentfont}
	\caption{SyncEdit for Lip Synchronization}\label{alg:lipsync}
	\KwIn{source video $X^{src}$, $\{t_i\}^{T}_{i=0}$, pretrained model $V_{\theta}$, prompt $c$, start step $n_{max}$, target audio $A^{tar}$, source audio $A^{src}$}
	\KwOut{target video $X^{tar}$}
	\KwInit{$X^{tar}_{t_{max}}=X^{src}_{t_{max}}=(1-t_{max})X^{src}+t_{max}\epsilon$}
	\For{$i = n_{max}$ \KwTo $0$}{
		$\epsilon_i \gets \mathcal{N}(0,I)$ \\
		\tcp{Compute source and target velocity}
		$V^{src}_{t_i} \gets V_{\theta}(X^{src}_{t_i}, A^{src}, t_i, c)$\\
		$V^{tar}_{t_i} \gets V_{\theta}(X^{tar}_{t_i}, A^{tar}, t_i, c)$\\
		\tcp{Annealed noise alignment}
		$\hat{\epsilon}_i \gets X^{src}_{t_i} + (1-t_i)V^{src}_{t_i}$\\
		$\tilde{\epsilon}_{t_{i-1}} \gets (1-t_{i-1})\hat{\epsilon}_i + t_{i-1}\epsilon_i$\\
		\tcp{Source and target sequence iteration}
		$X^{src}_{t_{i-1}} \gets (1-t_{i-1})X^{src} + t_{i-1}\tilde{\epsilon}_{t_{i-1}}$\\
		$X^{tar}_{t_{i-1}} \gets X^{tar}_{t_i} + \Delta t_{i} (V^{tar}_{t_i} - V^{src}_{t_i}) + X^{src}_{t_{i-1}} - X^{src}_{t_i}$
		
	}
	\KwReturn{$X^{tar}_0$}
\end{algorithm}

\section{SyncEdit}
\label{sec:synciedit}

Motivated by the aforementioned observations, we propose SyncEdit, a training-free framework for lip synchronization. An overview of the proposed framework is presented in Fig.~\ref{fig:overview}. Our approach reformulates the FlowEdit paradigm by replacing the conventional iterative editing trajectory with an iteration process defined directly on the target sequence, enabling an unbiased estimation of the desired output. Moreover, we introduce an annealed noise alignment strategy that progressively interpolates between stochastic Gaussian noise and noise estimated from the pre-trained diffusion model, resulting in a smoother and more stable editing trajectory.

In the following, we first introduce the proposed target-sequence iteration strategy and annealed noise alignment mechanism in detail. Based on these two components, we establish a unified training-free editing framework that performs accurate lip synchronization.

\begin{table*}[!t]
	\centering
	\caption{Quantitative results on HDTF Dataset.}
	\resizebox{\linewidth}{!}{
		\begin{tabular}{ccccccccc}
			\toprule
			\multirow{2}{*}{Methods}  & \multicolumn{3}{c}{Full Reference Metrics} & \multicolumn{3}{c}{No Reference Metrics} & \multicolumn{2}{c}{Lip Sync} \\
			\cmidrule(lr){2-4}
			\cmidrule(lr){5-7}
			\cmidrule(lr){8-9}
			& FID$\downarrow$ & FVD$\downarrow$ & CSIM$\uparrow$ & NIQE$\downarrow$ & BRISQUE$\downarrow$ & HyperIQA$\uparrow$ & LMD$\downarrow$ & LSE-C$\uparrow$ \\
			\midrule
			Wav2Lip\cite{PrajwalMNJ20wav2lip} & 14.912 & 543.340 & 0.852 & 6.495 & 53.372 & 45.822 & 10.007 & \underline{7.630} \\
			IP-LAP\cite{Zhong2023iplap} & 9.512 & 325.691 & 0.809 & 6.533 & 54.402 & 50.086 & 7.695 & 7.260 \\
			Diff2Lip\cite{Mukhopadhyay2024diff2lip} & 12.079 & 461.341 & 0.869 & 6.261 & 49.361 & 48.869 & 18.986 & 7.140 \\
			MuseTalk\cite{Zhang2024musetalk} & 8.759 & 231.418 & 0.862 & 5.824 & 46.003 & 55.397 & 8.701 & 6.890 \\
			LatentSync\cite{li2024latentsync} & 8.518 & 216.899 & 0.859 & 6.270 & 50.861 & 53.208 & 17.344 & \textbf{8.050} \\
			Omnisync\cite{peng2025omnisync} & \underline{7.855} & \underline{199.627} & 0.875 & \underline{5.481} & \underline{37.917} & \textbf{56.356} & \textbf{7.097} & 7.309 \\
			Ours(Humo1.7B) & 7.952 & 201.038 & \underline{0.879} & 5.604 & 39.527 & 54.714 & 7.698 & 7.157 \\
			Ours(Humo17B) & \textbf{7.623} & \textbf{190.299} & \textbf{0.884} & \textbf{5.385} & \textbf{37.412} & \underline{55.973} & \underline{7.182} & 7.286 \\
			\bottomrule
		\end{tabular}
	}
	\label{tab:exphdtf}
\end{table*}

\begin{table*}[!t]
	\centering
	\caption{Quantitative results on AIGC-LipSync Benchmark.}
	\resizebox{\linewidth}{!}{
		\begin{tabular}{ccccccccc}
			\toprule
			\multirow{3}{*}{Methods}  & \multicolumn{3}{c}{Full Reference Metrics} & \multicolumn{3}{c}{No Reference Metrics} & \multicolumn{2}{c}{Generation Success Rate} \\
			\cmidrule(lr){2-4}
			\cmidrule(lr){5-7}
			\cmidrule(lr){8-9}
			& FID$\downarrow$ & FVD$\downarrow$ & CSIM$\uparrow$ & NIQE$\downarrow$ & BRISQUE$\downarrow$ & HyperIQA$\uparrow$ & All Videos$\downarrow$ & \makecell{Stylized \\ Characters$\uparrow$} \\
			\midrule
			Wav2Lip\cite{PrajwalMNJ20wav2lip} & 22.989 & 562.245 & 0.727 & 5.392 & 42.816 & 50.511 & 71.38\% & 26.67\% \\
			IP-LAP\cite{Zhong2023iplap} & 14.686 & 247.402 & 0.796 & 5.546 & 45.153 & 53.174 & 45.53\% & 6.67\% \\
			Diff2Lip\cite{Mukhopadhyay2024diff2lip} & 23.542 & 403.149 & 0.692 & 5.440 & 42.442 & 50.335 & 74.63\% & 36.67\% \\
			MuseTalk\cite{Zhang2024musetalk} & 17.668 & 297.621 & 0.667 & 4.935 & 36.017 & 58.334 & 92.20\% & 67.78\% \\
			LatentSync\cite{li2024latentsync} & 15.374 & 263.111 & 0.751 & 5.342 & 41.917 & 54.648 & 74.96\% & 35.56\% \\
			Omnisync\cite{peng2025omnisync} & 10.681 & \underline{211.350} & 0.808 & \underline{4.588} & \underline{25.485} & \textbf{61.906} & \textbf{97.40\%} & \textbf{87.78\%} \\
			Ours(Humo1.7B) & \underline{10.516} & 213.136 & \underline{0.824} & 4.632 & 26.374 & 58.936 & 96.26\% & 79.98\% \\
			Ours(Humo17B) & \textbf{9.663} & \textbf{203.145} & \textbf{0.831} & \textbf{4.370} & \textbf{23.856} & \underline{61.448} & \underline{96.75\%} & \underline{84.27\%} \\
			\bottomrule
		\end{tabular}
	}
	\label{tab:explipsync}
\end{table*}

\subsection{Target Sequence Iteration}
\label{sec:targetseq}

We reframe the FlowEdit iteration process, transforming the iteration of the edit sequence into the iteration of the target sequence
\begin{align}
	X_{t_{max}}^{tar} &= X_{t_{max}}^{src} = (1-t_{max})X^{src} + t_{max} \epsilon, \\
	X_{t_i}^{tar} &= X_{t_i}^{edit} + X_{t_i}^{src} - X^{src}, \\
	X_{t_{i-1}}^{tar} &= X_{t_i}^{tar} + X_{t_{i-1}}^{edit} - X_{t_i}^{edit} + X_{t_{i-1}}^{src} - X_{t_i}^{src} \notag \\
	&= X_{t_i}^{tar} + \Delta t_i (V_{t_i}^{tar} - V_{t_i}^{src}) + X_{t_{i-1}}^{src} - X_{t_i}^{src}. \label{eq:tar}
\end{align}
$\{t_i\}_{i=0}^{T} $ is a monotonically decreasing time schedule. $X_{t_{max}}^{tar}$ and $X_{t_{max}}^{src}$ are initialized by adding the same Gaussian noise in $X^{src}$. $\Delta t_i = t_{i-1} - t_i$. The continuous-time ODE of the edit sequence in Eq.~\ref{eq:editode} is solved using a numerical ODE solver via time discretization. 

In contrast to the edit sequence, the target sequence yields an unbiased estimate of the desired target quantity. As illustrated in Fig.~\ref{fig:traj}, the key difference lies in the initialization of the trajectories. Specifically, the target trajectory is initialized as $X_{t_{max}}^{tar} = (1-t_{max})X^{src} + t_{max} \epsilon$. By comparison, the edit sequence is initialized directly from the clean source input, $X_{t_{max}}^{edit} = X^{src}$. The mismatch between the actual initialization and the theoretically required diffusion state introduces a systematic bias into the estimated editing trajectory, whereas the target sequence avoids this issue and yields an unbiased estimate of the desired target state.

By reformulating the iterative process in terms of the target sequence, we obtain a clearer interpretation of the underlying dynamical system and improve its analytical tractability. This reformulation not only facilitates theoretical analysis of the iteration dynamics but also provides a more principled foundation for the design of the associated noise parameterization.

\begin{figure*}[!t]
	\centering
	\includegraphics[width=\linewidth]{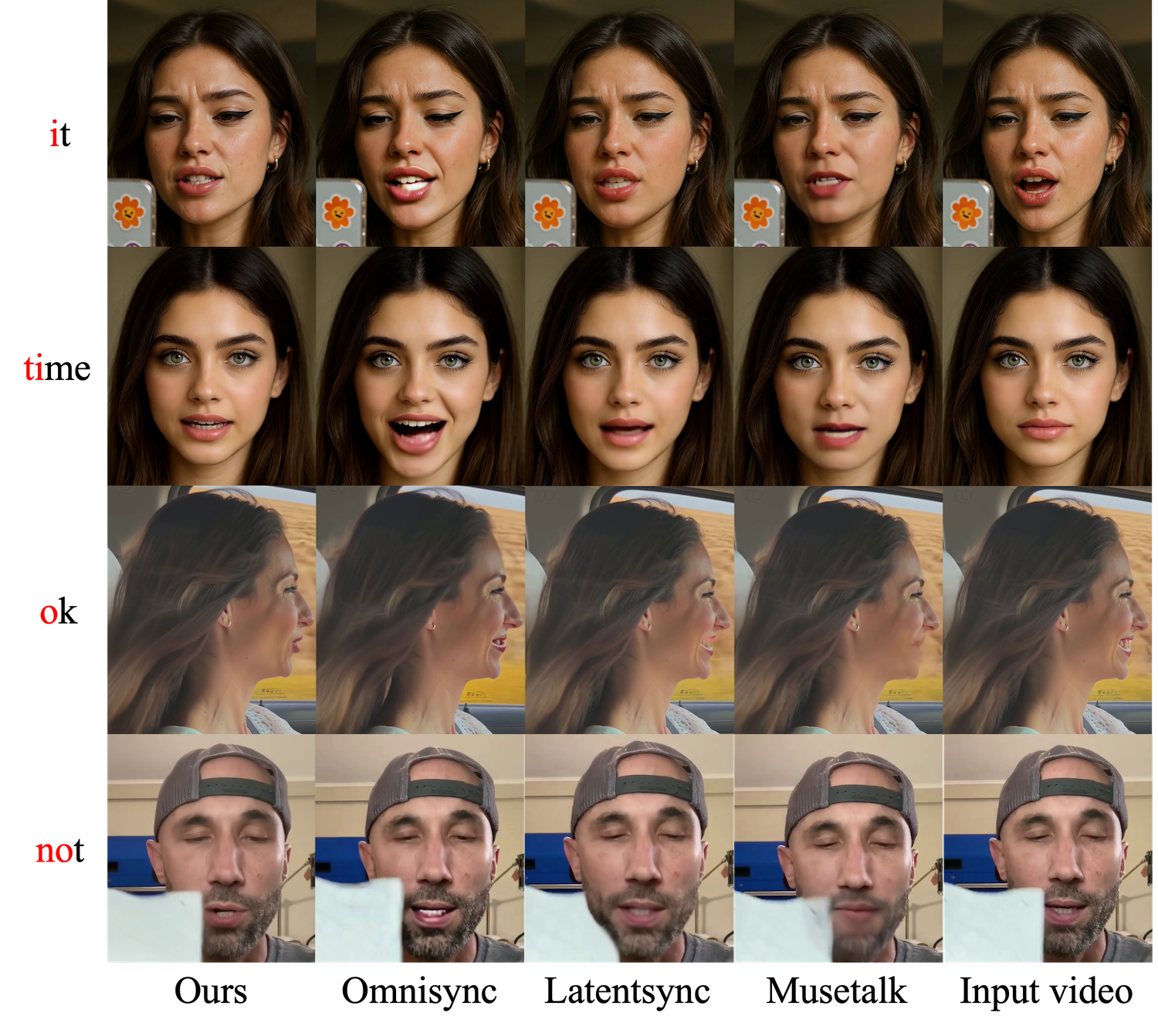}
	\caption{\textbf{Qualitative comparison with existing methods}. Compared with prior approaches, our method generates more accurate lip synchronization and more faithful dental structures while maintaining superior visual consistency. Even in challenging scenarios, including profile views and significant facial occlusions, our method produces natural and plausible lip movements with fewer visual artifacts.}
	\label{fig:lipsync}
\end{figure*}

\subsection{Annealed  Noise Alignment}

Eq.~\ref{eq:tar} can be further reformulated into the following form
\begin{equation}
	X_{t_{i-1}}^{tar} = X_{t_i}^{tar} + \Delta t_i V_{t_i}^{tar} + X_{t_{i-1}}^{src} - X_{t_i}^{src} +\Delta t_i V_{t_i}^{src},
\end{equation}
which can be interpreted as augmenting the iteration of the target sequence with the incremental variation induced by the source sequence iteration. This further indicates that the introduction of random noise into the source sequence in FlowEdit~\cite{kulikov2025flowedit} is suboptimal, since it leads to irregular trajectories and increases the iterative error of the source sequence.

As mentioned in DynaEdit~\cite{kulikov2026dynaedit}, the stochastic Gaussian noise sampled at each iteration introduces inconsistencies in the direction of the editing velocity, resulting in unstable editing trajectories and undesired artifacts. We introduce annealed noise alignment to replace the purely stochastic noise injection in FlowEdit. Instead of relying solely on randomly sampled Gaussian noise, $X^{src}_{t_{i-1}}$ is constructed from a weighted combination of Gaussian noise and diffusion-model-estimated noise, with the influence of Gaussian noise gradually annealed throughout the iterative process
\begin{align}
	\hat{\epsilon}_{t_i} = X^{src}_{t_i} &+ (1-t_i)V^{src}_{t_i}, \epsilon_{t_i} \sim \mathcal{N}(0,I) \\
	\tilde{\epsilon}_{t_{i-1}} &= (1- t_{i-1}) \hat{\epsilon}_{t_i} + t_{i-1} \epsilon_{t_i} \\
	X^{src}_{t_{i-1}} &= (1-t_{i-1})X^{src} + t_{i-1}\tilde{\epsilon}_{t_{i-1}}.
\end{align}
By adopting this modification, we obtain smoother iteration dynamics and alleviate error accumulation associated with random noise.

\subsection{Lip Synchronization}

The proposed method can be directly applied to pre-trained audio-to-video diffusion models to achieve accurate lip synchronization. By incorporating the aforementioned theoretical insights into the editing process, our approach effectively leverages the generalization capability and expressive power of pre-trained models, enabling precise alignment between speech audio and lip movements.

Algorithm~\ref{alg:lipsync} outlines the proposed SyncEdit procedure for audio-driven lip synchronization within a training-free generative flow framework. Given a source video $X^{src}$, a predefined time schedule $\{t_i\}^{T}_{i=0}$, and a pretrained audio-driven video generation diffusion model $V_{\theta}$, the algorithm synthesizes a target video $X^{tar}$ that preserves the visual identity and dynamics of the source while conforming to a target audio signal $A^{tar}$.

The algorithm begins by initializing the source and target trajectories from an identical noisy latent at the maximum time. At each timestep, the pretrained audio-driven diffusion model predicts the source and target velocity fields conditioned on the corresponding audio. Instead of propagating the editing trajectory as in FlowEdit, the source trajectory is first updated by estimating the underlying noise from the predicted source velocity and annealing it with Gaussian noise to construct the latent at the next timestep. The target trajectory is then obtained by integrating the relative editing velocity while following the updated source trajectory, yielding an unbiased estimate of the desired target sequence. By coupling target-sequence iteration with annealed noise alignment, the proposed algorithm constructs a smooth and stable editing trajectory, enabling accurate training-free lip synchronization using a pretrained audio-driven diffusion model.

\section{Experiments}
\label{sec:exp}

\begin{table}[!t]
	\centering
	\caption{\textbf{Ablation study} for our proposed method.}
	\resizebox{\linewidth}{!}{
		\begin{tabular}{cccccccc}
			\toprule
			Methods &  FID$\downarrow$ & FVD$\downarrow$ & CSIM$\uparrow$ & NIQE$\downarrow$ & BRISQUE$\downarrow$ & HyperIQA$\uparrow$ & LSE-C$\uparrow$ \\
			\midrule
			Edit sequence & 7.944 & 198.409 & 0.878 & 5.408 & 37.526 & 55.198 & 7.284 \\
			Random noise & 7.707 & 194.916 & 0.880 & 5.396 & 37.457 & 55.492 & 7.286 \\
			Ours & 7.623 & 190.299 & 0.884 & 5.385 & 37.412 & 55.973 & 7.286 \\
			\bottomrule
		\end{tabular}
	}
	\label{tab:ablation}
\end{table}

\subsection{Experimental Setups}

We utilize Humo~\cite{chen2025humo} for lip synchronization. Humo is a unified human-centric framework for video generation that enables audio-driven synthesis and facilitates collaborative control across multiple modalities. The total number of denoising steps is set to $20$, with an initial step of $4$. We evaluate lip synchronization using quantitative and qualitative metrics.

\subsection{Lip Synchronization}

We compare SyncEdit with OmniSync~\cite{peng2025omnisync}, LatentSync~\cite{li2024latentsync}, MuseTalk~\cite{Zhang2024musetalk}, IP-LAP~\cite{Zhong2023iplap}, Diff2Lip~\cite{Mukhopadhyay2024diff2lip}, ad Wav2Lip~\cite{PrajwalMNJ20wav2lip} on HDTF~\cite{zhang2021hdtf} dataset and AIGC-LipSync Benchmark~\cite{peng2025omnisync}. Since OmniSync is not publicly available, we use the qualitative results reported in its paper, while the quantitative results are obtained from the Kling website~\cite{kling}. We follow the same evaluation metrics and datasets as OmniSync. Visual quality is evaluated using FID (Fréchet Inception Distance), FVD (Fréchet Video Distance), and CSIM (Cosine Similarity). We additionally report no-reference perceptual metrics, including NIQE (Natural Image Quality Evaluator), BRISQUE (Blind/Referenceless Image Spatial Quality Evaluator), and HyperIQA~\cite{su2020hyperiqa}. Audio–visual synchronization performance is measured via LMD (Landmark Distance), and LSE-C (Lip Sync Error – Confidence). 

We evaluate Generation Success Rate (GSR) on the AIGC-LipSync benchmark for both the full set and stylized characters. The GSR quantifies the proportion of videos exhibiting correct lip synchronization as confirmed by human evaluators, providing a crucial metric for assessing the effectiveness of universal lip synchronization in AI-generated content.

\begin{figure}[!t]
	\centering
	\includegraphics[width=\linewidth]{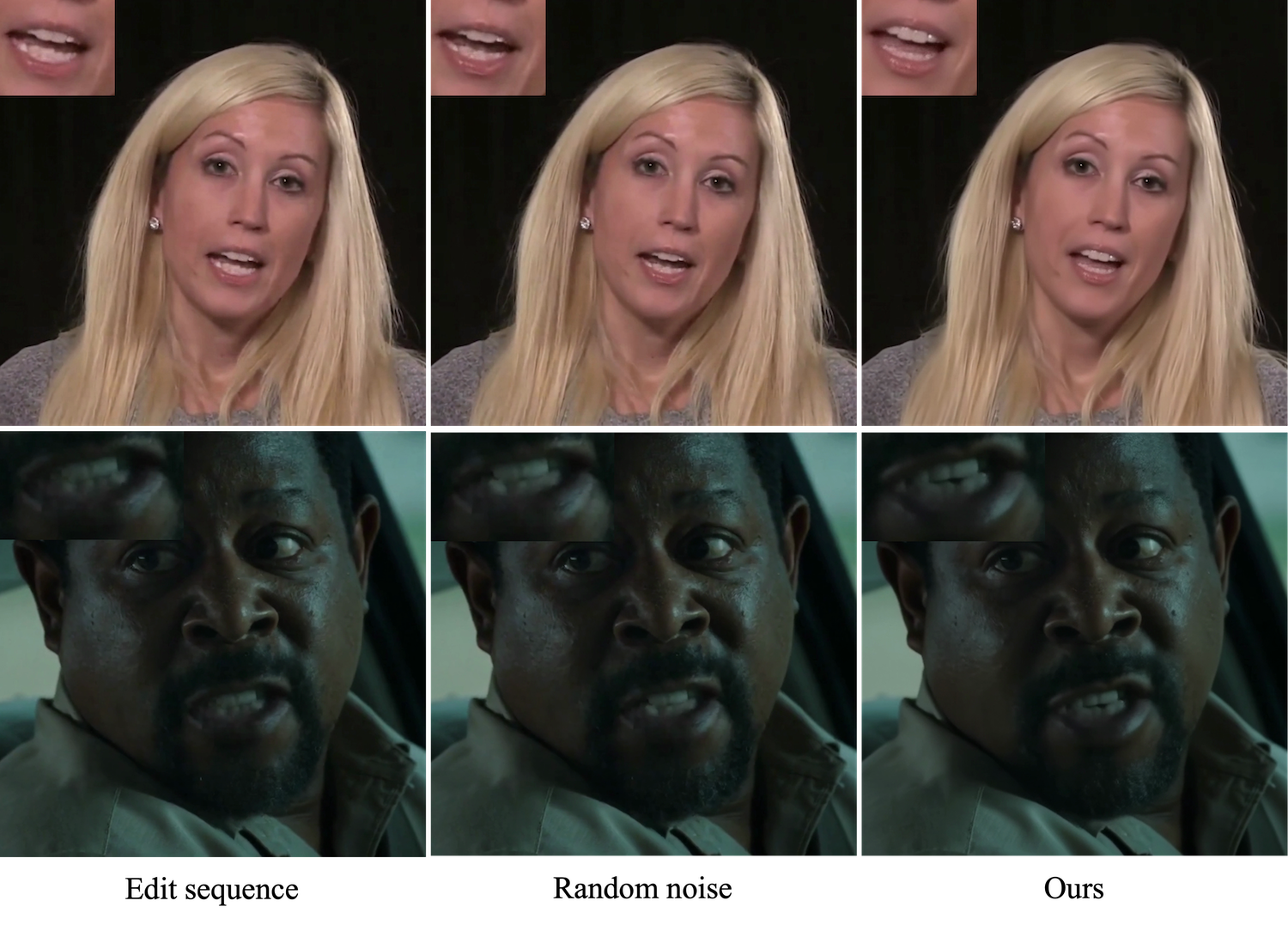}
	\caption{\textbf{Ablation study} for edit sequence and random noise. Edit sequence and stochastic noise injection tends to produce blurred dental details, whereas our method is capable of generating sharper and more clearly defined teeth.}
	\label{fig:ablation}
\end{figure}

The experimental results on the HDTF dataset are presented in Table~\ref{tab:exphdtf}. Our method achieves comparable performance to the current state-of-the-art approach, OmniSync. By formulating lip synchronization as an editing process, our approach minimally alters the original video while maximally preserving identity consistency, resulting in the lowest FID and FVD scores, as well as the highest CSIM. Regarding no-reference quality metrics, our method outperforms OmniSync in terms of NIQE and BRISQUE, while achieving slightly lower performance on HyperIQA. In terms of lip-synchronization-specific metrics, including LMD and LSE-C, our method performs slightly worse than OmniSync. LatentSync attains a highest LSE-C score, which can be attributed to its SyncNet-based loss.

The quantitative results on the AIGC-LipSync Benchmark are shown in Table~\ref{tab:explipsync}. Our method again achieves the lowest FID and FVD scores, together with the highest CSIM, demonstrating its strong capability in preserving visual quality and identity consistency. In terms of generation success rate, our approach performs slightly worse than OmniSync. This limitation can be attributed to Humo, which is primarily trained on human-centric datasets and thus exhibits a relatively lower driving success rate on stylized characters. Nevertheless, our method still outperforms other methods in this regard.

\begin{figure*}[!t]
	\centering
	\includegraphics[width=\linewidth]{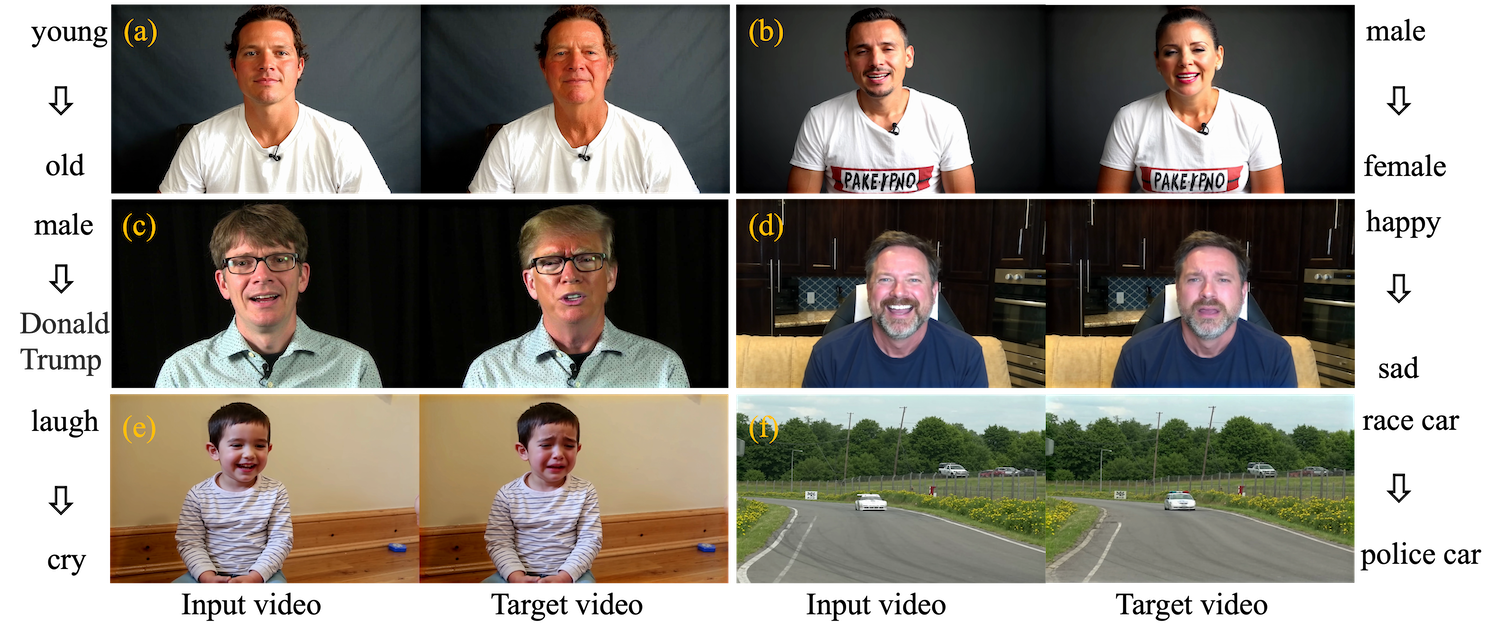}
	\caption{\textbf{Qualitative results of SyncEdit for Audio-visual Editing}. Our approach supports prompt-based manipulation of diverse attributes—including age (a), gender (b), person (c), emotion (d), behaviors (e), and even car categories (f), while jointly generating audio and video in a temporally synchronized and semantically consistent manner.}
	\label{fig:avedit}
\end{figure*}

The qualitative results are illustrated in Fig.~\ref{fig:lipsync}. The results of omnisync come from Kling's website~\cite{kling}. Our method consistently achieves more accurate lip synchronization while generating sharper and more faithful dental structures than competing approaches. Notably, in challenging cases involving large head poses (e.g., profile views) and facial occlusions, our method exhibits substantially stronger robustness, producing natural and visually consistent lip movements where other methods often suffer from noticeable artifacts or temporal inconsistencies. Comprehensive visual results are provided in the supplementary material.


\subsection{Ablation Study}

To validate the contribution of each component in our framework, we conduct an ablation study on key modules, including replacing the edit sequence with the target sequence and annealed noise alignment startegy. Quantitative results are reported in Table~\ref{tab:ablation}, and qualitative comparisons are presented in Fig.~\ref{fig:ablation}.

Although the iterative process over the edit sequence can produce noise-free results, it fails to yield an unbiased estimate of the target, which consequently degrades the fidelity and visual clarity of the generated results. In contrast, iterating over the target sequence more accurately approximates the underlying ODE, leading to improved generative quality as reflected by lower FID and FVD scores. As illustrated in Fig.~\ref{fig:ablation}, iterating over the edit sequence produces noticeably blurred teeth, whereas the target-sequence iteration yields significantly sharper and more clearly defined dental details.

Similarly, introducing randomly sampled Gaussian noise when computing the source sequence yields a non-smooth iterative trajectory, which adversely affects generation quality and leads to higher FID and FVD scores. In contrast, replacing stochastic noise with estimated noise produces a smoother trajectory, thereby improving the overall quality of the generated results. As shown in Fig.~\ref{fig:ablation}, incorporating random noise also results in degraded visual clarity, particularly in the dental region, while our approach preserves sharper structural details and achieves superior overall quality.

\subsection{Audio-Visual Editing}

Similarly, by leveraging a pre-trained audio–visual foundation model, SyncEdit extends naturally to prompt-driven audio–visual editing. Textual prompts serve as high-level semantic guidance, enabling coherent manipulation of both visual and auditory modalities.


We employ LTX-2~\cite{hacohen2026ltx2} for audio–visual editing. We conduct a qualitative evaluation of SyncEdit to assess its performance in this setting. As illustrated in Fig.~\ref{fig:avedit}, our proposed method demonstrates the ability to modify input videos according to given textual prompts, enabling controlled editing of attributes such as age, gender, identity, emotion, behaviors, and car categories. These results highlight the flexibility and generality of SyncEdit in handling diverse and semantically rich audio-visual editing tasks.



\section{Conclusion}
\label{sec:concl}

In this paper, we introduce SyncEdit, a training-free framework for lip synchronization. By reformulating the editing process and replacing the edit sequence with the target sequence in the FlowEdit paradigm, the proposed method enables an unbiased estimation of the desired output. Furthermore, by progressively aligning stochastic Gaussian noise with diffusion-model-estimated noise, we constructs a smoother and more stable editing trajectory, leading to improved stability and higher-quality generation. Extensive experiments demonstrate that our approach achieves strong performance and robustness across diverse scenarios, validating the effectiveness of the proposed design. We believe this training-free formulation provides a practical and efficient alternative to conventional fine-tuning-based editing methods.
{
    \small
    \bibliographystyle{ieeenat_fullname}
    \bibliography{main}
}


\end{document}